\documentclass[fleqn,10pt]{wlscirep}
\usepackage[utf8]{inputenc}
\usepackage[T1]{fontenc}
\usepackage{multirow} 
\usepackage{hhline} 
\usepackage{graphicx} 
\usepackage{xr}
\usepackage{graphicx}

\title{A personalized time-resolved 3D mesh generative model for unveiling normal heart dynamics}

\author[1,2,*]{Mengyun Qiao}
\author[3,4]{Kathryn A. McGurk}
\author[2,5]{Shuo Wang}
\author[1,6,7]{Paul M. Matthews}
\author[3]{Declan P. O'Regan}
\author[1,2,8]{Wenjia Bai}
\affil[1]{Department of Brain Sciences, Imperial College London, London, UK}
\affil[2]{Data Science Institute, Imperial College London, London, UK}
\affil[3]{MRC Laboratory of Medical Sciences, Imperial College London, London, UK}
\affil[4]{National Heart and Lung Institute, Imperial College London, London, UK}
\affil[5]{Digital Medical Research Center, School of Basic Medical Sciences, Fudan University and Shanghai Key Laboratory of MICCAI, Shanghai, China}
\affil[6]{UK Dementia Research Institute, Imperial College London, London, UK}
\affil[7]{Rosalind Franklin Institute, Harwell Science and Innovation Campus, Didcot, UK}
\affil[8]{Department of Computing, Imperial College London, London, UK}
\affil[*]{m.qiao21@imperial.ac.uk}

\begin{abstract}
Understanding the structure and motion of the heart is crucial for diagnosing and managing cardiovascular diseases, the leading cause of global death. There is wide variation in cardiac shape and motion patterns, that are influenced by demographic, anthropometric and disease factors. Unravelling the normal patterns of shape and motion, as well as understanding how each individual deviates from the norm, would facilitate accurate diagnosis and personalised treatment strategies. To this end, we developed a novel conditional generative model, MeshHeart, to learn the distribution of shape and motion patterns for the left and right ventricles of the heart. To model the high-dimensional and complex spatio-temporal mesh data, MeshHeart employs a geometric encoder to represent cardiac meshes in a latent space, followed by a temporal Transformer to model the motion dynamics of latent representations. Based on MeshHeart, we investigate the latent space of 3D+t cardiac mesh sequences and propose a novel distance metric termed latent delta, which quantifies the deviation of a real heart from its personalised normative pattern in the latent space. In experiments using a large cardiac magnetic resonance image dataset of 38,309 subjects from the UK Biobank, MeshHeart demonstrates a high performance in cardiac mesh sequence reconstruction and generation. Features defined in the latent space are highly discriminative for cardiac disease classification, whereas the latent delta exhibits strong correlations with clinical phenotypes in phenome-wide association studies. The code and the trained model will be released to benefit further research on digital heart modelling.
\end{abstract}
\begin{document}

\flushbottom
\maketitle
\thispagestyle{empty}

\section*{Introduction}
The heart is one of the most important and vital organs within the human body \cite{sanz2008imaging}. It is composed of four morphologically distinct chambers that function in a coordinated manner. The shape of the heart is governed by genetic and environmental factors \cite{aung2019genome, bonazzola2024unsupervised}, as well as a remodelling process observed in response to myocardial infarction, pressure overload, and cardiac diseases \cite{Meyer2020, Kim2018}. The motion of the heart follows a periodic non-linear pattern modulated by the underlying molecular, electrophysiological, and biophysical processes \cite{DBLP:journals/natmi/BelloDDBMHGWCRO19}. Unveiling the complex patterns of cardiac shape and motion will provide important insights for assessing the status of cardiac health in both clinical diagnosis and cardiovascular research \cite{puyol2017multimodal, Duchateau2020, bai2020population, corral2020digital}.

The current state-of-the-art for assessing cardiac shape and motion is to perform analyses of cardiac images, e.g., cardiac magnetic resonance (MR) images, and extract imaging-derived phenotypes (IDPs) of cardiac chambers \cite{schulz2020standardized,bai2020population}. Most imaging phenotypes, such as chamber volumes or ejection fractions, provide a global and simplistic measure of the complex 3D-temporal (3D+t) geometry of cardiac chambers \cite{hundley2022society,schulz2020standardized}. However, these global volumetric measures may not fully capture the dynamics and variations of cardiac function across individuals. Recent studies have shown that mesh-based cardiac shape and motion analyses can provide more detailed and clinically relevant insights \cite{gasparovici2024generative, piras2017morphologically, gilbert2019independent, mauger2022multi}. For example, Piras et al.\cite{piras2017morphologically} proposed to use spatio-temporal motion analysis to identify myocardial infarction. Gilbert et al.\cite{gilbert2019independent} highlighted stronger associations between cardiac risk factors and mesh-derived metrics in the UK Biobank dataset. Mauger et al.\cite{mauger2022multi} showed that mesh-based motion metrics could independently predict adverse cardiac events. This underscores the importance of establishing a precise computational model of cardiac status to define what a normal heart looks like and moves like. Nevertheless, it is a non-trivial task to describe the normative pattern of the 3D shape or even 3D-t motion of heart, due to the complexity in representing high-dimensional spatio-temporal data.

Recently, machine learning techniques have received increasing attention for cardiac shape and motion analysis \cite{DBLP:journals/natmi/BelloDDBMHGWCRO19,qi2020non,ye2023sequencemorph}. Most existing research focuses on developing discriminative machine learning models, that is, training a model to perform classification tasks between different shapes or motion patterns \cite{Suinesiaputra2017, Zheng2019, DBLP:journals/natmi/BelloDDBMHGWCRO19, Duchateau2020}. However, discriminative models offer only classification results and do not explicitly explain what the normative pattern of cardiac shape or motion looks like \cite{Kawel2020}. In contrast, generative machine learning models provide an alternative route. Generative models are capable of describing distributions of high-dimensional data, such as images \cite{gao2022get3d,xue2022giraffe,kim2023datid}, geometric shapes \cite{DBLP:conf/eccv/PetrovichBV22,DBLP:journals/corr/abs-2209-04066,DBLP:conf/iccv/PetrovichBV21}, or molecules \cite{swanson2024generative,jiang2024pocketflow}, which allow the representation of normative data patterns in the latent space of the model. In terms of generative modelling of the heart, recent developments focus on shape reconstruction and virtual population synthesis\cite{gasparovici2024generative, kong2024sdf4chd,wang2020deep,vukadinovic2023gancmri,gomez2022digital,muffoletto2023combining}. For example, Xia et al. proposed a method that integrates statistical shape priors with deep learning for four-chamber cardiac shape reconstruction from images \cite{xia2022automatic}. Gaggion et al. introduced HybridVNet, which combines CNNs with graph convolutions to perform shape reconstruction from multi-view images \cite{gaggion2023multi}. Dou et al. proposed a conditional flow-based variational autoencoder (VAE) for synthesising virtual populations of cardiac anatomy\cite{dou2023conditional}, and later developed a compositional generative model for multipart anatomical structures \cite{dou2024generative}. Beetz et al. introduced a variational mesh autoencoder that models population-wide variations in cardiac shapes with a hierarchical structure\cite{beetz2022interpretable}, and investigated the interpretability of the latent space extracted from a point cloud VAE \cite{beetz2021generating}. Although generative models have been explored for cardiac shape reconstruction \cite{gaggion2023multi,xia2022automatic}, shape modeling \cite{bonazzola2024unsupervised, beetz2022interpretable, dou2023conditional, dou2024generative}, image/video generation \cite{campello2022cardiac, qiao2023cheart, Hadrien2022} and data augmentation \cite{Gilbert2021}, their application to personalised normative modeling of the heart from population data remains underexplored.

Here, we provide the first endeavour to create a personalised normative model of 3D+t cardiac shape and motion, leveraging deep generative modelling techniques. Cardiac shape and motion are represented by a dynamic sequence of 3D surface meshes across a cardiac cycle. A novel geometric deep generative model, named MeshHeart, is developed to model the distribution of 3D+t cardiac mesh sequences. MeshHeart employs a graph convolutional network (GCN) \cite{DBLP:conf/iclr/KipfW17} to learn the latent features of the mesh geometry and a Transformer to learn the temporal dynamics of the latent features during cardiac motion. This integration enables MeshHeart to model the distributions across both spatial and temporal dimensions. MeshHeart functions as a conditional generative model, accounting for major clinical variables such as sex and age as the generation factor. This enables the model to describe personalised normative patterns, generating synthetic healthy cardiac mesh sequences for a specific patient or a specific subpopulation.

\begin{figure}[hbtp]
\centering
\includegraphics[width=\textwidth]{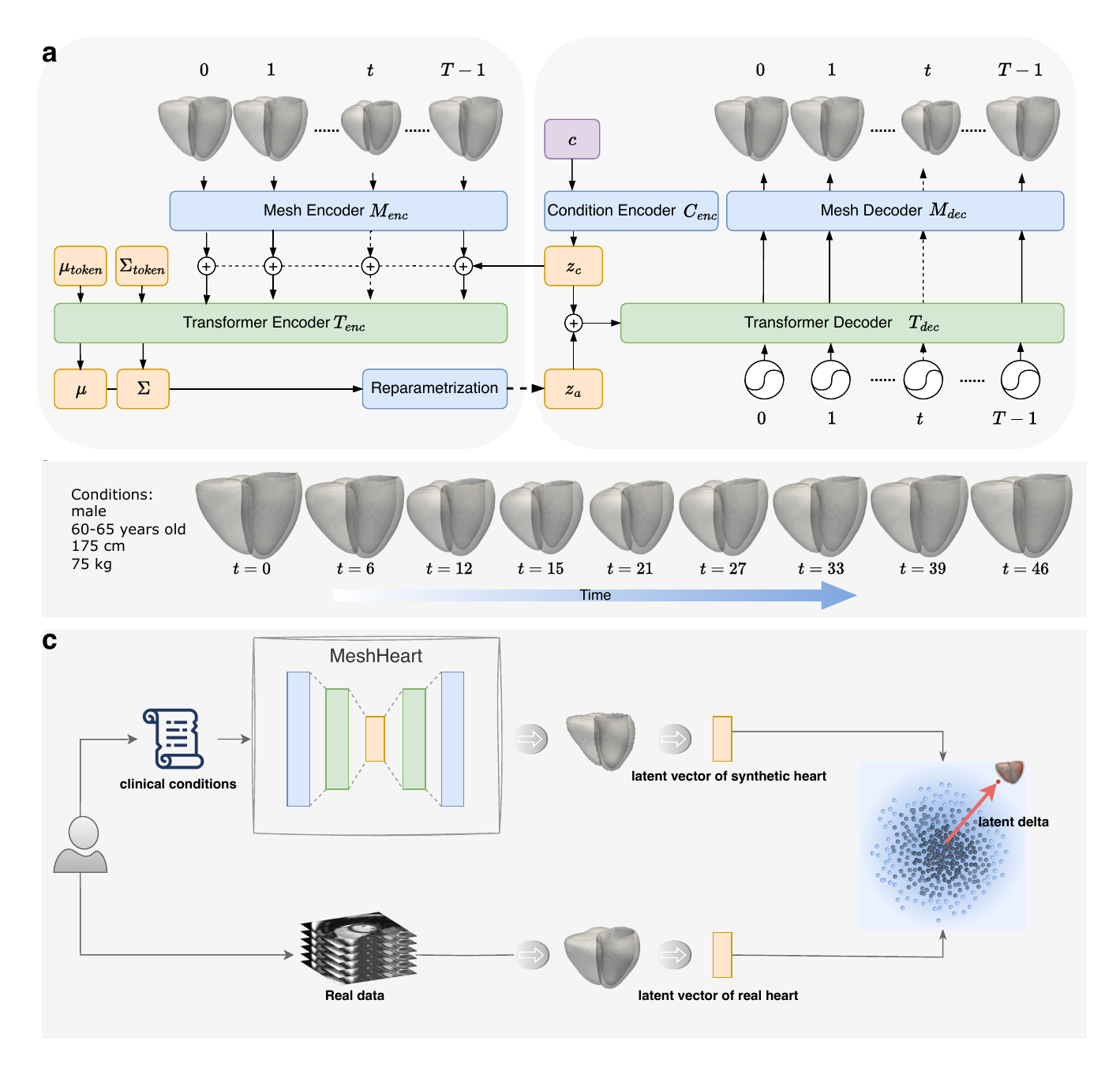}
\caption{{\bfseries An overview of the MeshHeart model.} {\bfseries a}, Model architecture: MeshHeart encodes a sequence of cardiac meshes using a mesh encoder $M_{\text{enc}}$ and encodes clinical factors using a conditional encoder $C_{\text{enc}}$. The encoder outputs across the time frames and along with distribution tokens $\mu_{token}$ and $\Sigma_{token}$, are processed by a temporal Transformer encoder $T_{\text{enc}}$. A Transformer decoder $T_{\text{dec}}$ and a mesh decoder $M_{\text{dec}}$ generate a 3D cardiac mesh sequence based on clinical factors. {\bfseries b}, Given a set of clinical factors, an example of the generated mesh sequence across time frames. {\bfseries c}, Conceptual framework: MeshHeart constructs a normative cardiac mesh sequence using personal information including age, sex, weight, and height. A real heart can be compared to its personalised norm by the latent vector. The latent delta $\Delta z$ is a distance defined between the latent vector of a synthetic normal heart (dark blue dot) and that of the real heart (red dot).} \label{fig_architecture}
\end{figure}

We train the proposed generative model, MeshHeart (Figure \ref{fig_architecture}(a)), on a large-scale population-level imaging dataset with 38,309 subjects from the UK Biobank \cite{bai2020population, petersen2015uk}. After training the model, for each individual heart, we can generate a personalised 3D+t cardiac mesh model that describes the normative pattern for this particular subpopulation that has the same clinical factors as the input heart, as illustrated in Figure \ref{fig_architecture}(c). In qualitative and quantitative experiments, we demonstrate that MeshHeart achieves high accuracy in generating the personalised heart model. Furthermore, we investigates the clinical relevance of the latent vector $z$ of the model and propose a novel distance metric (latent delta $\Delta z$), which measures the deviation of the input heart from its personalised normative pattern (Figure \ref{fig_architecture} (c)). We demonstrate that the latent vector and latent delta have a highly discriminative value for the disease classification task and they are associated with a range of clinical features in phenome-wide association studies.

\section*{Results}
\begin{figure}[btp]
\centering
\includegraphics[width=\textwidth]{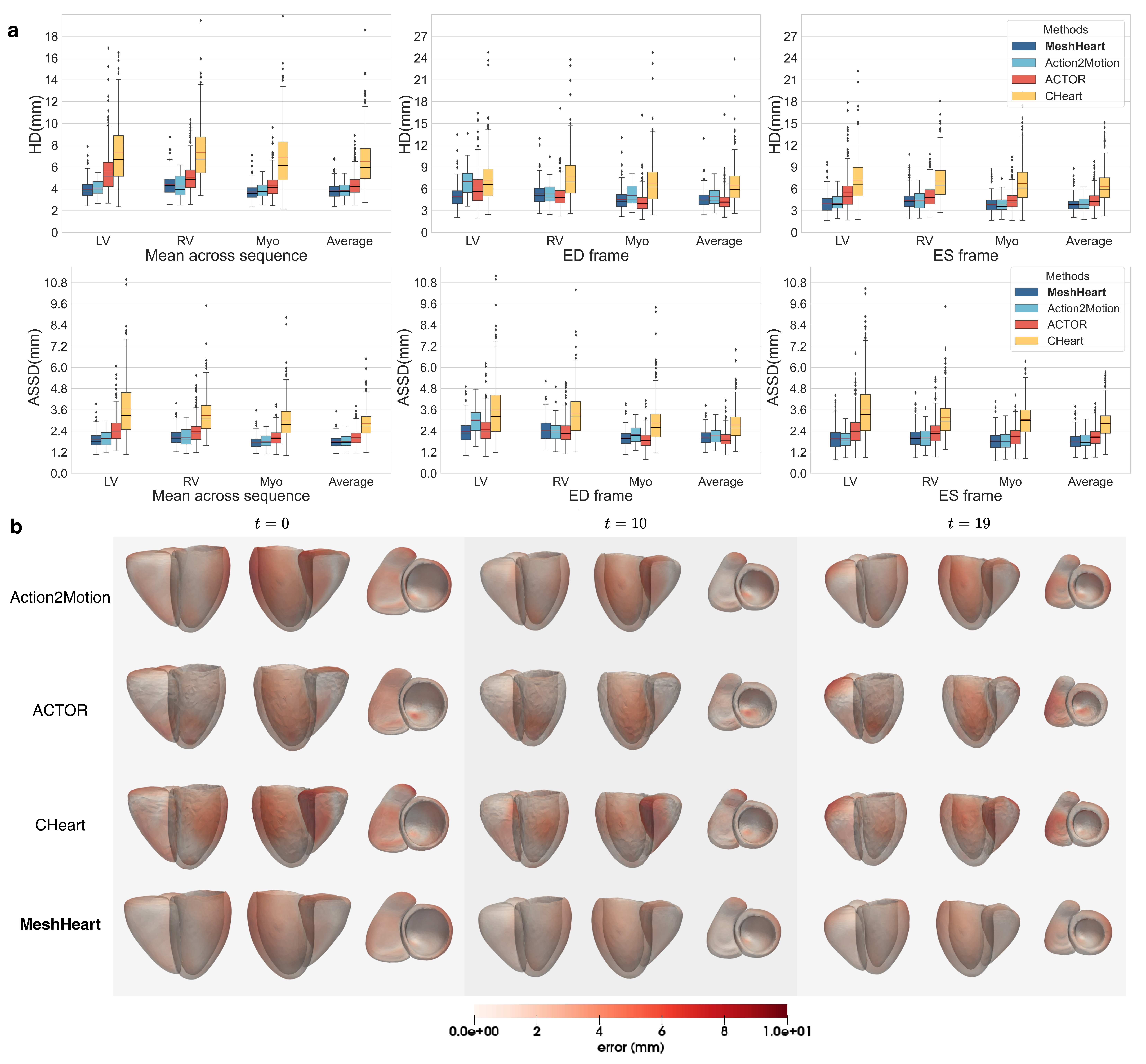}
\caption{{\bfseries Evaluation of the mesh reconstruction accuracy of MeshHeart, compared to three other methods; Action2Motion, ACTOR, and CHeart.} {\bfseries a}, Plots of the Hausdorff distance (HD) and average symmetric surface distance (ASSD) metrics. The metrics are calculated as the mean across all time frames, as well as at the end-diastolic frame (ED) and the end-systolic frame (ES). They are reported for the left ventricle (LV), the myocardium (Myo), the right ventricle (RV), and averaged across the anatomical structures. Lower values indicate better performance. {\bfseries b},  Visualisation of the reconstructed cardiac mesh sequence, coloured by the reconstruction error (in red) between the input mesh and reconstructed mesh. The mesh is visualised in three different imaging planes.} \label{fig_result1}
\end{figure}
\subsection*{MeshHeart learns spatial-temporal characteristics of the mesh sequence}
We first assessed the reconstruction capability of MeshHeart for 3D+t cardiac mesh sequences. The experiments used a dataset of 4,000 test subjects, with dataset detail described in Supplementary Table S1. Each input mesh sequence was encoded into latent representation and then decoded to reconstruct the mesh sequence. Reconstruction performance was evaluated using two metrics, the Hausdorff distance (HD) and the average symmetric surface distance (ASSD), which measure the difference between the input and reconstructed meshes. The HD metric quantifies the maximum distance between points in two sets, highlighting the maximum discrepancy between the original and reconstructed heart meshes. ASSD computes the average distance between the surfaces of two meshes, providing a more holistic evaluation of the model's accuracy. Evaluation was performed for three anatomical structures: the left ventricle (LV), the myocardium (Myo), and the right ventricle (RV). We compared the performance of MeshHeart to three baseline mesh generative models: Action2Motion \cite{DBLP:conf/mm/GuoZWZSDG020}, ACTOR \cite{DBLP:conf/iccv/PetrovichBV21}, and CHeart \cite{qiao2023cheart}.

Figure~\ref{fig_result1}(a) and Supplementary Table S2 report the reconstruction accuracy of MeshHeart, compared to other generative models. The metrics are reported as the average across all time frames, as well as at two representative time frames of cardiac motion: the end-diastolic frame (ED) and the end-systolic frame (ES). Overall, MeshHeart achieves the best reconstruction accuracy, outperforming other generative models, with the lowest HD of 4.163 mm and ASSD of 1.934 mm averaged across the time frames and across anatomical structures. Additionally, Figure~\ref{fig_result1}(b) visualises examples of the reconstructed meshes, with vertex-wise reconstruction errors overlaid, at different frames of the cardiac cycle ($t=0,10,19$ out of 50 frames in total). MeshHeart achieves lower reconstruction errors compared to the other models and maintains the smoothness of reconstructed meshes. We further conducted ablation studies to assess the contribution of each component to the model performance. These components are described in the Methods section and the detailed results are reported in Supplementary Table S5. Replacing GCN by linear layers results in an increased HD from $4.163~mm$ to $5.707~mm$, while replacing GCN by CNN results in a HD of $5.268~mm$, highlighting GCN's superiority in encoding mesh geometry. Substituting the Transformer with GRUs or LSTMs leads to an increased HD of $4.720~mm$ or $5.015~mm$, respectively, which demonstrates the advantage of using the Transformer for modelling long-range temporal dependencies. Other components such as the smoothness loss term and the distribution parameter tokens also contribute to the model performance. These results highlight MeshHeart's capability in learning spatial-temporal characteristics of cardiac mesh sequences.

\begin{figure}[btp]
\centering
\includegraphics[width=\textwidth]{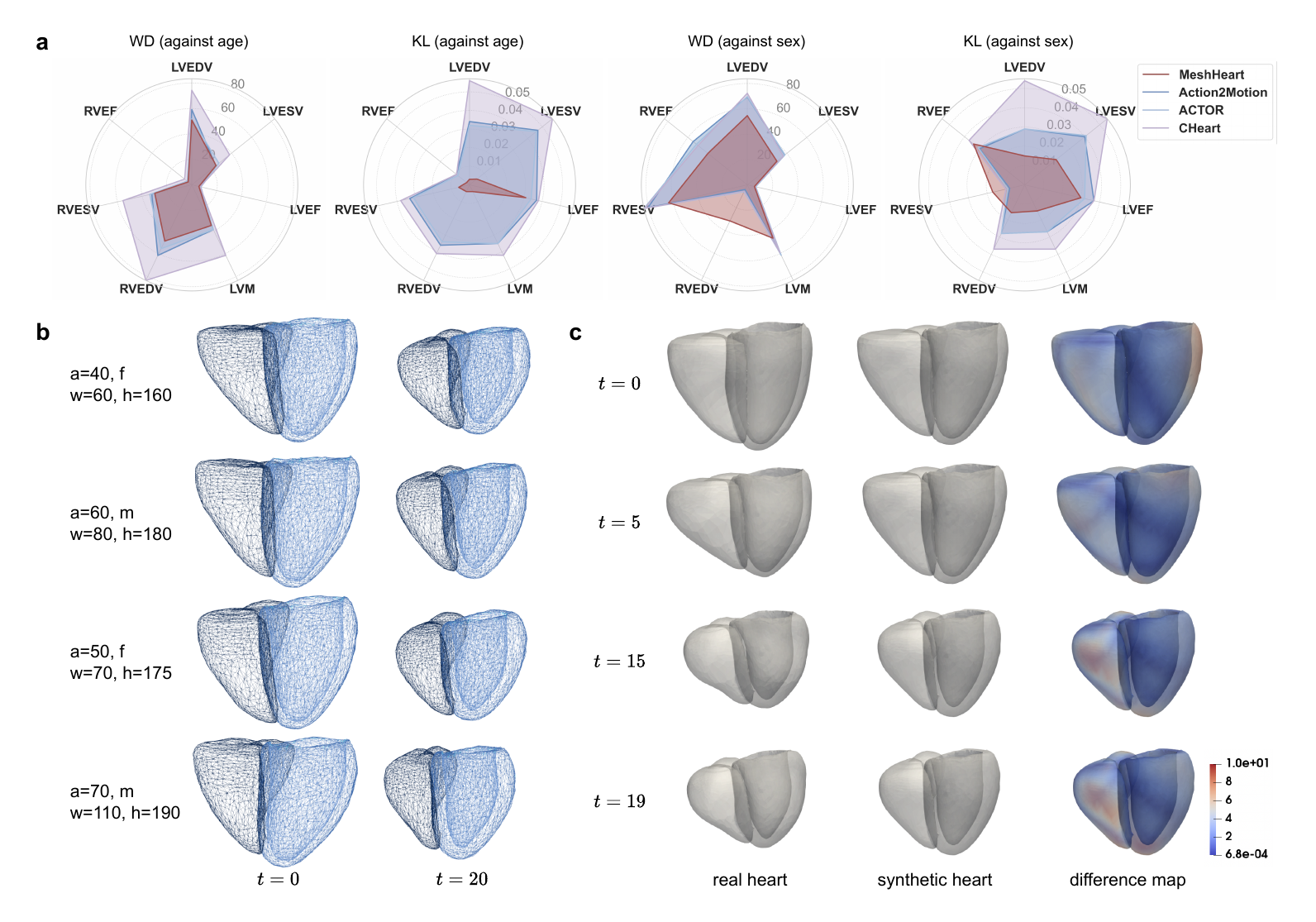}
\caption{{\bfseries Evaluation of the generation performance of MeshHeart.} \textbf{a}, Spider charts for the Wasserstein distance (WD) and the Kullback-Leibler(KL) divergence metrics, which quantify the distance between the generated and real data distributions. The data distribution is calculated as the histogram of a cardiac imaging phenotype (left ventricular end-diastolic volume (LVEDV), LV end-systolic volume (LVESV), LV ejection fraction (LVEF), LV myocardial mass (LVM), right ventricular end-diastolic volume (RVEDV), RV end-systolic volume (RVESV) and RV ejection fraction (RVEF)) against a clinical factor (age or sex). The metrics are plotted over a polar coordinate system, colour-coded by different methods. The smaller the metric (closer to the centre), the greater the similarity between the generated and real data distributions. \textbf{b}, Examples of generated 3D+t cardiac meshes with different generating factors, including age (a), sex (f/m for female/male), weight (w), and height (h). \textbf{c}, A side-by-side comparison of a real heart, the generated synthetic heart, and the difference map between them.} \label{figure_result2}
\end{figure}

\subsection*{MeshHeart resembles real data distribution}
Utilising the latent representations learnt by MeshHeart, we assessed the ability of the model to generate new synthetic cardiac mesh sequences that mimic real heart dynamics. To evaluate the fidelity and diversity of the generation, we calculated the similarity between the distributions of real meshes and generated synthetic meshes. For each real heart in the test set ($n = 4,000$), we applied MeshHeart to generate synthetic mesh sequences using the same clinical factors (age, sex, weight, and height) as the individual as the model input. During the generation stage, we chose 20 random samples from the Gaussian distribution of the latent space and generated the corresponding mesh sequences. For both real and synthetic meshes, clinically relevant metrics for cardiac structure and function were derived, including left ventricular end-diastolic volume (LVEDV), LV end-systolic volume (LVESV), LV ejection fraction (LVEF), LV myocardial mass (LVM), right ventricular end-diastolic volume (RVEDV), RV end-systolic volume (RVESV) and RV ejection fraction (RVEF). For each metric $m$, its probability distributions against age $P(m|c=age)$ and against sex $P(m|c=sex)$, were calculated. The similarity between real and synthetic probability distributions was quantified using the Kullback-Leibler (KL) divergence \cite{cover1999elements} and the Wasserstein distance (WD) \cite{arjovsky2017wasserstein}, with a lower value denoting a higher similarity, i.e. better generation performance. KL divergence is a metric from information theory that evaluates the dissimilarity between two probability mass functions. Similarly, WD measures the dissimilarity between two probability distributions. MeshHeart's ability to replicate real data distributions is quantitatively demonstrated in Figure~\ref{figure_result2}(a). MeshHeart achieves lower KL and WD scores compared to existing methods, as shown by radar plots with the smallest area, suggesting that the synthetic data generated by the proposed model closely aligns with the real data distribution for clinically relevant metrics. Supplementary Tables S3 and S4 report the detailed KL divergence and WD scores for different methods.

For qualitative assessment, Figure~\ref{figure_result2}(b) presents four instances of synthetic cardiac mesh sequences for different personal factors (age, sex, weight, and height). For brevity, only two frames ($t=0, 20$) are shown. The figure demonstrates that MeshHeart can mimic authentic cardiac movements, showing contractions across time from diastole to systole. Figure~\ref{figure_result2}(c) compares a real heart with a synthetic normal heart, at different time frames ($t=0, 5, 15, 19$), demonstrating the capability of MeshHeart in replicating both the real cardiac structure as well as typical motion patterns.

We also examined the latent representation learnt by MeshHeart using t-SNE (t-distributed Stochastic Neighbour Embedding) visualisation \cite{van2008visualizing} as illustrated in Supplementary Figure S1. The t-SNE plot projects the 64-dimensional latent representation of a mesh, extracted from the last hidden layer of the Transformer Encoder $T_{\text{enc}}$, onto a two-dimensional space, with each point denoting a mesh. It shows 10 sample sequences. For each sample, the latent representations of the meshes across time frames form a circular pattern that resembles the rhythmic beating of the heart \cite{fukuta2008cardiac}. 

\begin{figure}[btp]
\centering
\includegraphics[width=\textwidth]{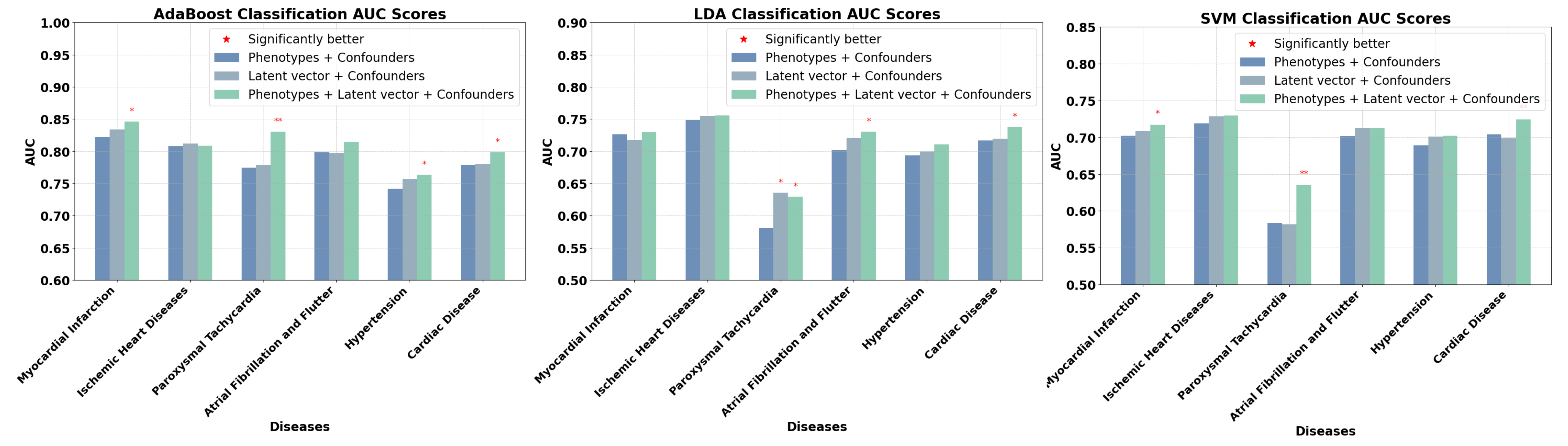}
\caption{{\bfseries Comparison of disease classification performance, in terms of the AUC (area under the curve) scores, when different feature sets are used.} The three feature sets being compared include "Phenotypes $+$ Confounders (Age, Sex, Weight, Height)", "Latent vector $+$ Confounders", and "Phenotypes $+$ Latent vector $+$ Confounders", with each feature set represented by a unique color in the plot. The three subplots show the performance of three different classifiers, AdaBoost, LDA, and SVM. The x-axis denotes the disease type and the y-axis denotes the AUC score. For the three feature sets, each feature set is compared to one of the other two features sets using the DeLong's test. A single asterisk (*) denotes a notable difference ($p < 0.05$), indicating that a feature set outperforms another feature set significantly, while a double asterisk (**) indicates that a feature set outperforms both of the other two feature sets significantly.} \label{figure_result3}
\end{figure}

\subsection*{MeshHeart latent vector improves cardiovascular disease classification performance}
\begin{figure}
\centering
\includegraphics[width=\textwidth]{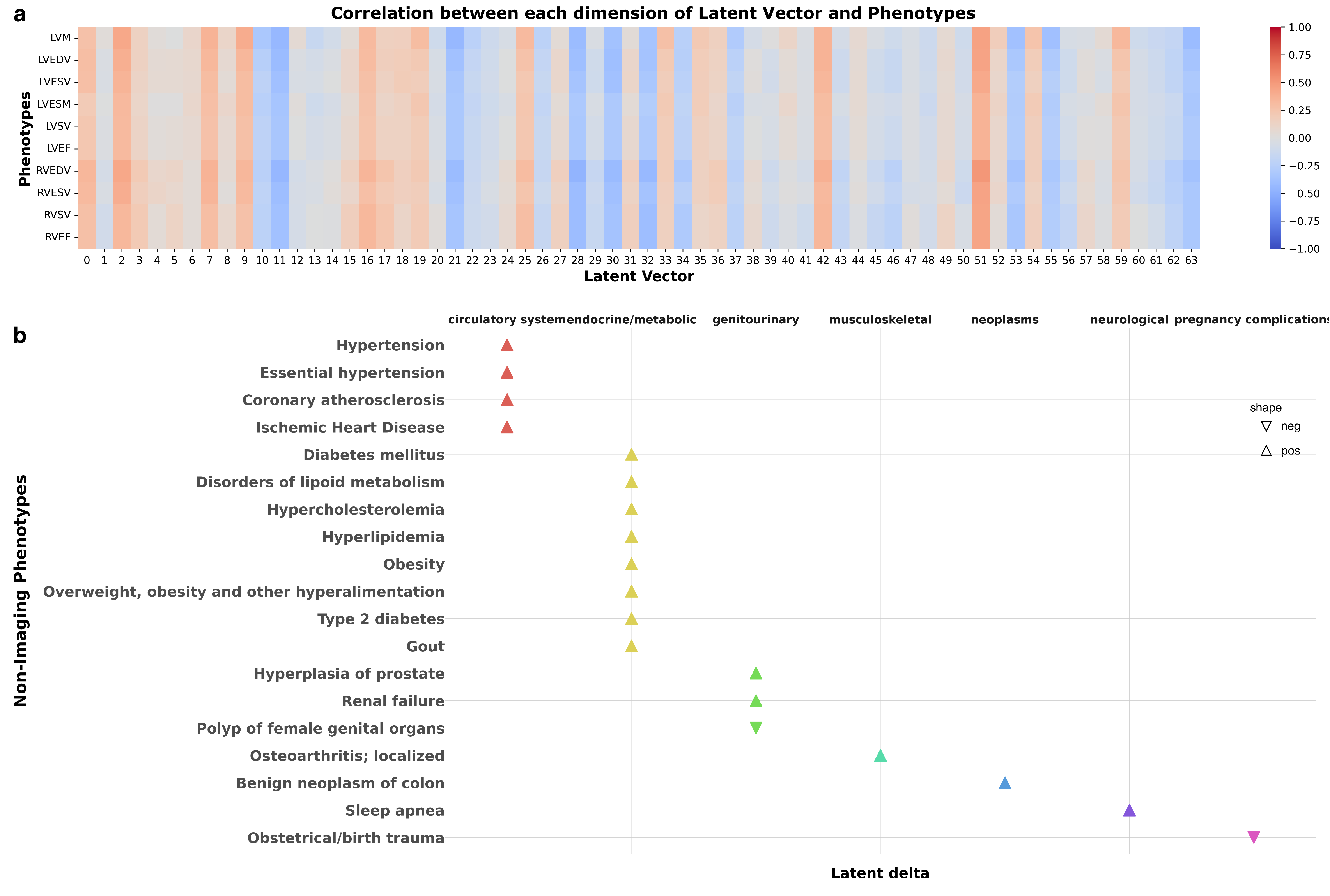}
\caption{{\bfseries Association studies for the latent vector and the latent delta with imaging-derived phenotypes and clinical features.} \textbf{a}, Heatmap of the Pearson correlation coefficients between imaging phenotypes and the 64-dimensional latent vector. The intensity of the colour reflects the magnitude and direction of the correlation, where blue signifies negative correlations and red signifies positive correlations. Darker shades indicate a stronger correlation between the vector and the phenotypes. \textbf{b}, Phenome-wide association study (PheWAS) between the latent delta $\Delta z$ and unbiased categories of clinical features. The y-axis lists the clinical outcomes where a signficiant association was identified. The x-axis uses different colors to represent different disease categories. Each triangle denotes a significant PheWAS association, adjusted for multiple comparisons using the Bonferroni correction for 1,163 clinical features analysed. These clinical features encompass both clinical outcomes (e.g., diseases and diagnoses) and phenotypes, covering a wide range of characteristics and measurements. The shape of each triangle indicates the direction of the effect. This analysis included 17,000 participants.}\label{figure_result4}
\end{figure}

After demonstrating the generative capability of MeshHeart, we explore its potential for clinical applications, in particular using its latent space which provides a low-dimensional representation of cardiac shape and motion. The latent feature analyses were conducted on 17,309 subjects. More than half (58.5\%) had a reported diagnosis of at least one disease. We employ the latent vector $z$ of each mesh sequence, a 64-dimensional vector, as the feature for correlation analysis and for cardiac disease classification. Figure \ref{figure_result4}(a) shows that the latent vector exhibits strong correlations with conventional imaging phenotypes, such as LVM, LVEDV, and RVEDV etc. Figure~\ref{figure_result3} and Supplementary Table S6 compare the classification performance of six cardiac diseases when using different feature sets. The three evaluated feature sets include "Phenotypes + Confounders (Age, Sex, Weight, Height)", "Latent Vector + Confounders", and "Phenotypes + Latent Vector + Confounders". The classification performance is evaluated using the AUC (area under the curve) scores for three different classifiers: AdaBoost, linear discriminant analysis (LDA), and support vector machine (SVM). The six cardiovascular diseases include myocardial infarction (ICD-10 code I21), ischemic heart diseases (I24), paroxysmal tachycardia (I47), atrial fibrillation and flutter (I48), hypertension (I10), and cardiac disease (I51). Figure~\ref{figure_result3} shows that using imaging phenotypes alone led to moderate AUC scores (e.g., 0.8361 and 0.8201 for myocardial infarction and ischemic heart diseases using with AdaBoost). Using the latent vector resulted in increased AUC scores (0.8557 and 0.8453). Combining both imaging phenotypes and the latent vector further improved the AUC scores (0.8762 and 0.8472), indicating the usefulness of the latent vector for cardiovascular disease classification. These results demonstrate the model's ability to discriminate not only between normal and abnormal cardiac states, but also among specific disease conditions.

For the AdaBoost classifier, using feature sets comprising the latent vector, as well as the combination of phenotypes and the latent vector, consistently outperformed the performance of the phenotypes set alone (e.g., 0.8291 and 0.8316 for cardiac disease using latent vector and combined feature sets), implying that incorporating the latent vector improved the classification accuracy. The trend was particularly noticeable for myocardial infarction, hypertension, and cardiac diseases, where the combined phenotypes and latent vector feature set substantially improved the AUC scores (0.8762, 0.7738 and 0.8316 for myocardial infarction, hypertension and cardiac disease). While the model was trained using a normal healthy heart dataset, it has learnt a rich latent representation to encode diverse shape and motion patterns for different subpopulations in this large dataset. The resulting latent vector captures deviations in the latent space that are indicative of specific disease outcomes, as demonstrated by the experimental results. The LDA and SVM classifiers demonstrated that of the three feature sets, the combined phenotypes and latent vector feature set achieved the highest AUC scores (e.g., 0.6728 and 0.6479 for hypertension with LDA and SVM). However, for certain diseases such as ischemic heart disease, classifiers using only phenotypes  (e.g., 0.7381 and 0.7123 for ischemic heart diseases with LDA and SVM) outperformed those that used only the latent vector (0.7277 and 0.6975), but still fell short of their combination (0.7492 and 0.7214). Overall, the results show that integrating imaging phenotypes, the latent vector along with confounders provides the best discriminative feature set for classification.

\subsection*{Latent delta for phenome-wide association studies (PheWAS)}
For each individual heart, we use MeshHeart to generate a normal synthetic heart using the same clinical factors as this individual. This synthetic heart can be regarded as a personalised normative model learnt from a specific subpopulation. We define the latent delta $\Delta z$ to be the difference between the latent vectors of an individual heart and its personalised norm, quantified using the Euclidean distance. The latent delta characterises the deviation of the shape and motion patterns of an individual heart from the normal pattern for a subpopulation with the same clinical factors (Figure \ref{fig_architecture}(c)). A phenome-wide association study (PheWAS) was performed to explore the clinical relevance of $\Delta z$, as shown in Figure~\ref{figure_result4}(b). The PheWAS revealed significant associations between the latent delta $\Delta z$ and an unbiased set of clinical outcomes, including circulatory system diseases, endocrine/metabolic diseases, genitourinary diseases, musculoskeletal diseases and neoplasms.

The latent delta has been shown to correlate with phenotypes such as LVM and LVEF (Figure \ref{figure_result4}(a)), which serve as indicators of cardiac structure and function. Conditions such as hypertension, lipid/cholesterol abnormalities, and diabetes can induce changes in these cardiac phenotypes. For example, hypertension likely results in an increased LVM and may be linked to a reduced LVEF due to the heart's adaptation to prolonged high blood pressure. In a similar vein, diabetes can exert metabolic stress on the heart, which can lead to changes in cardiac volume and ejection fraction. These modifications in the structure and motion patterns of the heart, as captured by the latent delta, provide a mechanistic explanation for the associations observed in the PheWAS results. 

In Figure 5(b), the direction of effect shows the relationship between $\Delta z$ and the clinical outcome. A positive effect indicates that an increase in $\Delta z$ is associated with a higher probability of the outcome. In contrast, a negative effect indicates that a higher $\Delta z$ reduces the likelihood of the outcome. For example, a negative effect for birth trauma suggests that a higher $\Delta z$ is associated with a reduced likelihood of birth trauma. These directional effects provide insight into how deviations in cardiac structure and function relate to specific clinical outcomes, highlighting potential associations for further in-depth clinical investigation.


\section*{Discussion}
This work contributes to the growing field of generative artificial intelligence for science, with a specific application in cardiac imaging. The proposed MeshHeart model is a new generative model that can facilitate improved understanding of the complexities of 3D+t cardiac shape and motion. In this study, we made four major contributions. First, we developed MeshHeart using a dataset of 38,309 subjects from a large UK population \cite{petersen2015uk}, capturing the variation in cardiac structures and clinical characteristics. Second, we demonstrated MeshHeart's capability to generate a \textit{normal} heart, accounting for clinical factors such as age, sex, weight, and height. This established a personalised normative model for cardiac anatomy. Third, we investigated the latent vector of MeshHeart and demonstrated its associations with conventional imaging phenotypes and usefulness for enhancing disease classification performance. Finally, we propose a novel latent delta ($\Delta z$) metric. This metric provides a way for quantifying the difference between an individual heart and the normative model, as well as for investigating the associations between the spatial-temporal characteristics of the heart and various health outcomes.

MeshHeart's reconstruction capability was assessed using HD and ASSD metrics. Using these two metrics, we compared the model with other models along with an ablation study. Employing geometric convolutions and a temporal Transformer, the model reconstructed more accurate cardiac mesh sequences compared to the other state-of-the-art models. This is is due to the reason that geometric convolutions are proficient in encoding mesh geometry, and the Transformer is effective in capturing long-range temporal dependencies. The ablation study confirms the essential role of geometric convolutions and the temporal Transformer in increasing the performance of the model, as detailed in the Supplementary Table S5. We also compared MeshHeart against a previous work CHeart \cite{qiao2023cheart}. CHeart employs segmentation as a representation method for the cardiac structure, whereas MeshHeart employs the mesh representation. The results show that mesh provides a powerful representation for modelling the 3D geometry as well for tracking temporal motion, as it essentially allows monitoring the movement of each individual point over time. 

The generative capabilities of MeshHeart, as illustrated by the results in Figure~\ref{figure_result2} and Supplementary Tables S3 and S4, demonstrate its proficiency as a generative model, able to replicate a 'normal' heart based on certain clinical factors including demographics (age, sex) and anthropometrics (weight, height). These four factors have shown strong correlations with heart structure and function across various individuals \cite{nethononda2015gender,heckbert2006traditional,bai2020population}. They form a reliable basis for constructing a normal heart model for an individual, as shown in Figure~\ref{figure_result2}(b). Our analysis in Figure~\ref{figure_result2}(a) and Supplementary Tables S3 and S4 focused on age and sex, employing WD and KL divergence to assess the similarity between the real and synthetic data distributions. Lower WD and KL metrics suggest that MeshHeart effectively represents demographic diversity, making the synthetic data beneficial for potential clinical and research purposes. The incorporation of additional clinical variables in the future, such as blood pressure and medical history, could improve the representation of cardiac health and diseases, thus enabling more potential applications for downstream tasks.

The latent vector obtained from the MeshHeart demonstrated its discriminative power for disease classification tasks. Incorporating the latent vector as feature substantially improves the classification accuracy for a range of cardiovascular conditions, as illustrated in Figure~\ref{figure_result3}. Although conventional imaging phenotypes can also be used as a feature set for the classification model, their classification performance was surpassed by the augmented feature set that also includes the latent vector, suggesting that the latent vector may contain some information not provided by the imaging phenotypes. Combining imaging phenotypes with the latent vector and confounders consistently achieved the best classification performance, regardless of the classification model used, demonstrating the benefit of integrating multiple data sources to represent the status of the heart. Some dimensions of the latent vector exhibit high correlations with conventional cardiac phenotypes, which are essential for assessing cardiovascular disease risk. The high correlation with the latent vector underscores their clinical analysis potential.

PheWAS uses a data-driven approach to uncover unbiased associations between cardiac deviations and disease diagnoses. Our analysis found that greater deviations in heart function are linked to increased risks of endocrine/metabolic and circulatory diseases. These cardiac diseases suggest underlying metabolic problems such as insulin resistance or metabolic disturbances observed in diabetes and obesity, which affect the structure and performance of the heart \cite{ortega2016obesity,ormazabal2018association}. Likewise, they indicate wider circulatory conditions such as hypertension and atherosclerosis, which can lead to heart failure and ischemic heart disease\cite{ford2012ideal}. Understanding these relationships is crucial for risk stratification, personalised medicine, and prevention strategies, highlighting the need for thorough cardiac evaluations in clinical management \cite{binu2017heart}.

Although this work advances the science in personalised cardiac modelling, there are several limitations. First, the personalised normative model relies on a restricted range of generating factors, including age, sex, weight, and height, as we aim to develop a standard healthy heart. Including additional elements in the future such as diseases or environmental factors such as air pollution and noise\cite{bhatnagar2017environmental} could improve our understanding of their impacts on cardiac anatomy and function. Second, the model uses a cross-sectional dataset from the UK Biobank for both training and testing purposes. However, it does not include a benchmark for the progression of cardiac aging, which could be addressed by employing a longitudinal dataset to evaluate the model. Repeated scans are expected in the near future from the UK Biobank. Third, this study focuses on modelling the dynamic mesh sequence to describe cardiac shape and motion. It does not aim to model the underlying electrophysiology or biomechanics of the heart, which are also essential for cardiac modelling and understanding cardiac function \cite{mann2005mechanisms,trayanova2011whole}. Additionally, the explainability of latent vectors could be explored, as understanding the specific information each latent dimension captures is crucial for clinical interpretation and validation. Finally, our method does not incorporate long-axis images, which limits its ability to capture the mitral, tricuspid, or aortic valves for assessing valvular function. Mauger et al.\cite{mauger2019right} used 2-chamber and 4-chamber long-axis images to identify tricuspid and mitral valve points, so that the motion of the valve points can be tracked and modelled using principal component analysis.

In conclusion, this study presents MeshHeart, a generative model for cardiac shape modelling. By training and evaluating the model on a population-level dataset from the UK Biobank, we demonstrate that MeshHeart not only achieves a high reconstruction accuracy but also excels in generating synthetic cardiac mesh sequences that closely resemble the real heart. The latent vector of the generative model and the novel latent delta metric provide new avenues of research to improve disease classification and personalised healthcare. These findings pave the way for future research on cardiac modelling and may inspire the development of generative modelling techniques for other types of biomedical data. 

\section*{Methods}\label{sec5}
\subsection*{Generative model architecture}\label{model}
Figure~\ref{fig_architecture}(a) illustrates the architecture of the proposed generative model, MeshHeart. Given a set of clinical conditions $c$, our goal is to develop a model that can generate a dynamic 3D cardiac mesh sequence, $X_{0:T-1} = \{x_0, x_1, \cdots, x_{T-1}\}$,  where $T$ denotes the number of time frames, that corresponds to the conditions $c$. Figure~\ref{fig_architecture}(b) shows an example of the input conditions and the generated mesh sequence. Without losing generality, we take age, sex, weight and height as conditions $c$ in this work. Age, weight and height are continuous variables, whereas sex is a binary variable. Each cardiac mesh $x_t = (v_t, e_t)$ is a graph with a set of vertices $v$ and a set of edges $e$ connecting them. 

The proposed generative model consists of a mesh encoder $M_{\text{enc}}$, a Transformer encoder $T_{\text{enc}}$, a condition encoder $C_{\text{enc}}$, a Transformer decoder $T_{\text{dec}}$, and a mesh decoder $M_{\text{dec}}$. These components are designed to work together to learn the probability distribution $p_\theta(x|z_c)$ of the cardiac mesh sequence conditioned on clinical attributes, where $\theta$ represents the decoder parameters, and $z_c$ denotes the condition latent vector. The condition encoder $C_{\text{enc}}$, implemented as a multi-layer perceptron (MLP), maps the clinical conditions $c$ into a condition latent vector $z_c$. 

The mesh encoder $M_{\text{enc}}$, implemented as a graph convolutional network (GCN), processes the input cardiac mesh sequence $x_{0:T-1}$. It extracts latent representations $z_{0:T-1}$, where each vector $z_t$ corresponds to a latent representation of the cardiac mesh at time frame $t$. These latent vectors serve as intermediate representations of the cardiac mesh sequence.

The latent vectors $z_{0:T-1}$ from the mesh encoder are concatenated with the condition latent vector $z_c$ to form a sequence of input tokens to the Transformer encoder $T_{\text{enc}}$. The Transformer encoder $T_{\text{enc}}$ captures temporal dependencies across the sequence, which comprises $L$ layers of alternating blocks of multi-head self-attention (MSA) and MLP. To ensure stability and effective learning, LayerNorm (LN) is applied before each block and residual connections are applied after each block. Similar to the class token in the vision Transformer \cite{Dosovitskiy2021}, we append the input tokens $z_{0:T-1}$ with two learnable parameters $\mu_{token}$ and $\Sigma_{token}$, named as \textit{distribution parameter tokens}, which parameterize a Gaussian distribution over the latent space.  In the Transformer output layer, we extract the outputs from the distribution parameter tokens as distribution parameters $\mu$ and $\Sigma$. We then use the reparameterisation trick \cite{DBLP:journals/corr/KingmaW13} to derive the latent $z_a$ from $\mu$ and $\Sigma$, as shown in Figure~\ref{fig_architecture}(a). The encoding process is formulated as,
\begin{equation}
\label{eq:za}
  \begin{aligned}[b]
        &z_{input} =[\mu _{token};\Sigma_{token} ;z_0;z_1;...;z_{T-1}] \\
        &z'^l =MSA(LN(z^{l-1})) + z^{l-1},l=1,..,L \\
        &z^l =LN\left [MLP(LN(z'^l) \right ] \\
        &z_a =\mu+\epsilon \cdot \Sigma, \epsilon \sim \mathcal {N}(0,\mathbf{1})
  \end{aligned}
\end{equation}

The resulting latent vector $z_a$, derived after the reparameterisation step, captures the information about the distribution of the mesh sequence. This vector is concatenated with the condition latent vector $z_c$ to form the input to the Transformer decoder $T_{\text{dec}}$. The decoder uses these concatenated vectors as keys and values in the self-attention layer, while sinusoidal temporal positional encodings \cite{Dosovitskiy2021} serve as queries to incorporate temporal information.  The temporal positional encoding $p_t$ at time frame $t$ is defined using the sinusoidal function with the same dimension $d$ as $z_a$:
\begin{equation}
    {p_t}^{(i)}= \begin{cases}\sin \left(t/10000^{2i/d}\right), & \text { if } i=2 k \\ \cos \left(t/10000^{2i/d}\right), & \text { if } i=2 k+1\end{cases}
\end{equation}
where $i$ denotes the dimension index. The Transformer decoder outputs a sequence of latent vectors, each corresponding to a mesh representation at a time point of the cardiac cycle. The latent vectors generated by the Transformer decoder are passed through the mesh decoder $M_{\text{dec}}$, composed of fully connected layers, to reconstruct the 3D+t cardiac mesh sequence $X'_{0:T-1}$.

\subsection*{Probabilistic modelling and optimisation}
Following the VAE formulation\cite{DBLP:journals/corr/KingmaW13,DBLP:conf/iclr/HigginsMPBGBML17}, we assume a prior distribution $p(z_a)$ over the latent variable $z_a$. The prior $p(z_a)$, together with the decoder (constructed by $T_\text{dec}$ and $M_\text{dec}$), defines the joint distribution $p(x, z_a | z_c)$. To train the model and perform inference, we need to compute the posterior distribution $p(z_a|x,z_c)$, which is generally intractable. To turn the intractable posterior inference problem $p(z_a|x,z_c)$ into a tractable problem, we introduce a parametric encoder model (constructed by $C_{\text{enc}}, M_{\text{enc}}, T_{\text{enc}}$) $q_{\phi}(z_a|x,z_c)$ with $\phi$ to be the variational parameters, which approximates the true but intractable posterior distribution $p(z_a|x,z_c)$ of the generative model, given an input $x$ and conditions $c$:
\begin{equation}
q_{\phi}(z_a|x,z_c)\approx p_{\theta}(z_a|x,z_c)
\label{eq:optimize}
\end{equation}
where $q_{\phi}(z_a|x,z_c)$ often adopts a simpler form, e.g. the Gaussian distribution \cite{DBLP:journals/corr/KingmaW13,DBLP:conf/iclr/HigginsMPBGBML17}. By introducing the approximate posterior $q_{\phi}(z_a|x,z_c)$, the log-likelihood of the conditional distribution $p_{\theta}(x|z_c)$ for input data $x$, also known as evidence, can be formulated as:
\begin{equation}
\begin{aligned}
\log{p_{\theta}(x|z_c)}
&= \mathbb{E}_{z_a \sim q_{\phi}(z_a|x,z_c)}\log\left[{p_{\theta}(x|z_c)} \right] \\
&= \mathbb{E}_{z_a \sim q_{\phi}(z_a|x,z_c)}\log\left[{\frac{p_{\theta}(x,z_a|z_c)}{q_{\phi}(z_a|x,z_c)}}\right]
+ \mathbb{E}_{z_a \sim q_{\phi}(z_a|x,z_c)}\log\left[\frac{q_{\phi}(z_a|x,z_c)}{p_{\theta}(z_a|x,z_c)}\right]
\end{aligned}
\label{eq:logp}
\end{equation}
where the second term denotes the Kullback-Leibler (KL) divergence $D_{KL}(q_{\phi}\parallel p_{\theta})$, between $q_{\phi}(z_a|x,z_c)$ and $p_{\theta}(z_a|x,z_c)$ \cite{DBLP:journals/corr/KingmaW13,DBLP:conf/iclr/HigginsMPBGBML17}. It is non-negative and zero only if the approximate posterior $q_{\phi}(z_a|x,z_c)$ equals the true posterior distribution $p_{\theta}(z_a|x,z_c)$. Due to the non-negativity of the KL divergence, the first term in Eq.~\ref{eq:logp} is the lower bound of the evidence $\log[p_{\theta}(x|z_c)]$, known as the evidence lower bound (ELBO). Instead of optimising the evidence $\log[p_{\theta}(x|z_c)]$ which is often intractable, we optimise the ELBO:
\begin{equation}
\min_{\theta, \phi} ELBO = -\log[p_{\theta}(x|z_c)] + D_{KL}
\label{eq:ELBO}
\end{equation}

\subsection*{Training loss function}
Based on the ELBO, we define the concrete training loss function, which combines the mesh reconstruction loss $\mathcal{L}_R$, the Kullback-Leibler (KL) loss $\mathcal{L}_{KL}$, and a mesh smoothing loss $\mathcal{L}_S$. The mesh reconstruction loss $\mathcal{L}_R$ is defined as the Chamfer distance between the reconstructed mesh sequence $X'_{0:T-1}=(V',E')$ and the ground truth $X_{0:T-1}=(V,E)$, formulated as $\mathcal{L}_R = \frac{1}{T}\sum_{t=0}^{T-1} D_{cham}(V'_t,V_t)$, where $D_{cham}$ denotes the Chamber distance \cite{fan2017point}, $V'_t$ and $V_t$ denote the mesh vertex coordinates for the reconstruction and the ground truth respectively: 
\begin{equation}
    D_{cham}(V_t,V'_t)=\frac{1}{\left |V_t  \right |}\sum_{v_t\in V_t}\min_{v'_t\in V'_t}\left \| v_t-v'_t \right \|_2+\frac{1}{\left |V'_t  \right |}\sum_{v'_t\in V'_t}\min_{v_t\in V_t}\left \| v'_t-v_t \right \|_2
\end{equation}
In the VAE, the distribution of the latent space for $z_a$ is encouraged to be close to a prior Gaussian distribution. The KL divergence is defined between the latent distribution and the Gaussian prior distribution. To control the trade-off between distribution fitting and diversity, we adopt the $\beta$-VAE formulation \cite{DBLP:conf/iclr/HigginsMPBGBML17}. The KL loss $\mathcal{L}_{KL}$ is formulated as
\begin{equation}
    \mathcal{L}_{KL} = \beta \cdot KL(\mathcal{N}(\mu, \Sigma) \parallel \mathcal{N}(0, \mathbf{I}))
\end{equation}
which encourages the latent space $\mathcal {N}(\mu,\Sigma)$ to be close to the prior Gaussian distribution $\mathcal {N}(0,\mathbf{I})$. 

The Laplacian smoothing loss penalises the difference between neighbouring vertices such as sharp changes on the mesh \cite{DBLP:conf/graphite/NealenISA06, desbrun1999implicit}. It is defined as:
\begin{equation}
    \begin{aligned}[b]
    & \mathcal{L}_{S} = \frac{1}{T}\sum_{t=0}^{T-1} D_{smooth}(V'_t,E'_t) \\
    &D_{smooth}(V,E) =\sum_{v_i\in V}\frac{1}{|V|}\Big\|\sum_{j\in N_i}\frac{1}{|N_i|}(v_j-v_i)\Big\|_2
    \end{aligned}
\end{equation}
where $N_i$ denotes the neighbouring vertices adjacent to $v_i$. The total loss function $L$ is a weighted sum of the three loss terms:
\begin{equation}
\mathcal{L} = \mathcal{L}_R + \mathcal{L}_{KL} + \lambda_s \cdot \mathcal{L}_{S}
\end{equation}

In terms of implementation, the mesh encoder $M_{\text{enc}}$ has three GCN layers and one fully connected (FC) layer. The mesh decoder $M_{\text{dec}}$ is composed of five FC layers. The Transformer encoder $T_{\text{enc}}$ and decoder $T_{\text{dec}}$ consist of two layers, four attention heads, a feed-forward size of 1,024, and a dropout rate of 0.1. The latent vector dimensions for the mesh and condition were set to 64 and 32, respectively. The model contains approximately 69.71 million parameters and was trained on an NVIDIA RTX A6000 GPU (48 GB) using the Adam optimiser with a fixed learning rate of $10^{-4}$ for $300$ epochs. Training was performed with a batch size of one cardiac mesh sequence, consisting of 50 time frames. The cardiac mesh at each time frame consists of 22,043 vertices and 43,840 faces. The weights $\beta$ and $\lambda_s$ in the loss function were empirically set to 0.01 and 1.

\subsection*{Personalised normative model, latent vector, and latent delta}
MeshHeart is trained on a large population of asymptomatic hearts. Once trained, it can be used as a personalised normative model to generate a synthetic mesh sequence of a normal heart with certain attributes $c$, including age, sex, weight, and height. For each real heart, we can then compare the real cardiac mesh sequence to the synthetic normal mesh sequence of the same attributes, to understand the deviation of the real heart from its personalised normative pattern. 

To represent a cardiac mesh sequence in a low-dimensional latent space, we extract a latent vector after the Transformer encoder $T_{\text{enc}}$ but before the reparameterisation step. The latent vector is calculated as the mean of the latent vectors at the Transformer encoder output layer across 50 time frames. For calculating the latent delta, we quantify the deviation of the latent vector of the real heart to the latent vector of a group of synthetic hearts of the same attributes. Given conditions $c$, 100 samples of the latent variable $z_a$ are drawn from a standard Gaussian distribution, $z_a \sim \mathcal{N}(\mathbf{0}, \mathbf{I})$, where $z_a$ denotes the latent space after reparameterisation in the VAE formulation. Each sample $z_a$ is concatenated with the condition latent vector $z_c$ and passed through the Transformer decoder and mesh decoder to generate a synthetic cardiac mesh sequence. Following synthetic mesh generation, each synthetic mesh sequence is provided to the mesh encoder $M_{\text{enc}}$ and Transformer encoder $T_{\text{enc}}$, to generate latent vectors across 50 time frames at the Transformer output later, subsequently averaged to form the latent vector $z^{synth}$. The real heart mesh sequence is provided to the mesh encoder $M_{\text{enc}}$ and Transformer encoder $T_{\text{enc}}$ for calculating the latent vector $z^{real}$ in the same manner.

With the latent vector $z^{real}$ for the real heart and the latent vector $z^{synth}$ for the synthetic heart, we define the latent vector as the Euclidean distance between $z^{real}$ and $z^{synth}$. As we draw 100 synthetic samples to represent a subpopulation with the same attributes, the latent delta $\Delta z$ is defined as:
\begin{equation}
    \Delta z = \left\| z^{\text{real}} - \frac{1}{100} \sum_{i=1}^{100} z^{\text{synth}}_i \right\|_2
\end{equation}
where $i$ denotes the sample index. The latent delta $\Delta z$ provides a robust metric to evaluate individual differences in cardiac structure and motion, quantifying the deviation of the real heart from its personalised normative model.

\subsection*{Data and experiments}
This study used a data set of 38,309 participants obtained from the UK Biobank \cite{petersen2015uk}. Each participant underwent cine cardiac magnetic resonance (CMR) imaging scans. From the cine CMR images, a 3D mesh sequence is derived to describe the shape and motion of the heart. The mesh sequence covers three anatomical structures, LV, Myo and RV. Each sequence contains 50 time frames over the course of a cardiac cycle. To derive cardiac meshes from the CMR images, automated segmentation \cite{bai2018automated} was applied to the images. The resulting segmentations were enhanced using an atlas-base approach \cite{8624549}, by registering multiple high-resolution cardiac atlases \cite{bai2015bi,8624549} onto the segmentations followed by label fusion, resulting in high-resolution segmentations. A 3D template mesh \cite{bai2015bi} was then fitted to the high-resolution segmentations at the ED and ES frames using non-rigid image registration, generating ED and ES cardiac meshes. Subsequently, motion tracking was performed using Deepali \cite{Schuh2024-si}, a GPU-accelerated version of the non-rigid registration toolbox MIRTK \cite{rueckert1999nonrigid}, on cardiac segmentations across the time frames. Deformation fields were derived using a free-form deformation (FFD) model with a control point spacing of \([8, 8, 8]\). The registration objective function included Dice similarity as the primary similarity metric and B-spline bending energy regularisation with a weight of 0.01. The deformation fields were derived between time frames and applied to propagate the ED mesh and ES mesh across the cardiac cycle. The proposed meshes were averaged using weighted interpolation based on temporal proximity to ED and ES \cite{bai2020population} to ensure temporal smoothness of the resulting mesh sequence. All cardiac meshes maintained the same geometric structure.

The dataset was partitioned into training/validation/test sets for developing the MeshHeart model and a clinical analysis set for evaluating its performance for disease classification task. Briefly, MeshHeart was trained on 15,000 healthy subjects from the Cheadle imaging centre. For parameter tuning and performance evaluation, MeshHeart was evaluated on a validation set of 2,000 and a test set of 4,000 healthy subjects, from three different sites, Cheadle, Reading, and Newcastle centres. For clinical analysis, including performing the disease classification study and latent delta PheWAS, we used a separate set of 17,309 subjects from the three imaging centres, including 7,178 healthy subjects and 10,131 subjects with cardiac diseases and hypertension. PheWAS was undertaken using the PheWAS R package with clinical outcomes and coded phenotypes converted to 1,163 categorical PheCodes. P-values were deemed significant with Bonferroni adjustment for the number of PheCodes. The details of the dataset split and the definition of disease code are described in Supplementary Table S1.

\subsection*{Method comparison}
To the best of our knowledge, there is no previous work on generative modelling for 3D+t cardiac mesh sequences. To compare the generation performance of MeshHeart, we adapt three state-of-the-art generative models originally proposed for other tasks: 1) Action2Motion \cite{DBLP:conf/mm/GuoZWZSDG020}, originally developed for human motion generation; 2) ACTOR \cite{DBLP:conf/iccv/PetrovichBV21}, developed for human pose and motion generation; 3) CHeart\cite{qiao2023cheart}, developed for the generation of cardiac segmentation maps, instead of cardiac meshes. We modified these models to adapt to the cardiac mesh generation task.
\section*{Data availability statement}
The raw imaging data and non-imaging participant characteristics are available from UK Biobank to approved researchers via a standard application process at http://www.ukbiobank.ac.uk/register-apply. Access the data at UK Biobank(https://www.ukbiobank.ac.uk/).
\section*{Code availability statement}
The code for this research is available on GitHub (https://github.com/MengyunQ/MeshHeart).

\section*{Acknowledgements}
This research was conducted using the UK Biobank Resource under Application Number 18545. Images were reproduced with kind permission of UK Biobank. We wish to thank all UK Biobank participants and staff. We also thank Weitong Zhang for helpful discussions on the methodology. This work is supported by EPSRC DeepGeM Grant (EP/W01842X/1) and the BHF New Horizons Grant (NH/F/23/70013). K.A.M. is supported by the British Heart Foundation (FS/IPBSRF/22/27059, RE/18/4/34215) and the NIHR Imperial College Biomedical Research Centre. S.W. is supported by Shanghai Sailing Program (22YF1409300), CCF-Baidu Open Fund (CCF-BAIDU 202316) and International Science and Technology Cooperation Program under the 2023 Shanghai Action Plan for Science (23410710400). P.M.M. acknowledges generous personal support from the Edmond J. Safra Foundation and Lily Safra, an NIHR Senior Investigator Award and the UK Dementia Research Institute, which is funded predominantly by UKRI Medical Research Council. D.P.O is supported by the Medical Research Council (MC\_UP\_1605/13); National Institute for Health Research (NIHR) Imperial College Biomedical Research Centre; and the British Heart Foundation (RG/19/6/34387, RE/24/130023, CH/P/23/80008). 

\section*{Author contributions statement}
MY.Q. and WJ.B. conceived the study. MY.Q. conducted the experiments 2-5, K.A.M. conducted the experiment 5b, MY.Q. and WJ.B. analysed the results, D.P.O. and P.M.M. provided data resources. All authors reviewed the manuscript. 

\section*{Competing interests}

D.P.O has consulted for Bayer AG and Bristol Myers-Squibb. K.A.M. has consulted for Checkpoint Capital LP. None of these activities are directly related to the work presented here. P.M.M. has received consultancy or speaker fees from Roche, Merck, Biogen, Rejuveron, Sangamo, Nodthera, and Novartis. P.M.M. has received research or educational funds from Biogen, Novartis, Merck, and GlaxoSmithKline. All other authors have nothing to disclose.

\begin{thebibliography}{10}
\urlstyle{rm}
\expandafter\ifx\csname url\endcsname\relax
  \def\url#1{\texttt{#1}}\fi
\expandafter\ifx\csname urlprefix\endcsname\relax\def\urlprefix{URL }\fi
\expandafter\ifx\csname doiprefix\endcsname\relax\def\doiprefix{DOI: }\fi
\providecommand{\bibinfo}[2]{#2}
\providecommand{\eprint}[2][]{\url{#2}}

\bibitem{sanz2008imaging}
\bibinfo{author}{Sanz, J.} \& \bibinfo{author}{Fayad, Z.~A.}
\newblock \bibinfo{journal}{\bibinfo{title}{Imaging of atherosclerotic cardiovascular disease}}.
\newblock {\emph{\JournalTitle{Nature}}} \textbf{\bibinfo{volume}{451}}, \bibinfo{pages}{953--957} (\bibinfo{year}{2008}).

\bibitem{aung2019genome}
\bibinfo{author}{Aung, N.} \emph{et~al.}
\newblock \bibinfo{journal}{\bibinfo{title}{Genome-wide analysis of left ventricular image-derived phenotypes identifies fourteen loci associated with cardiac morphogenesis and heart failure development}}.
\newblock {\emph{\JournalTitle{Circulation}}} \textbf{\bibinfo{volume}{140}}, \bibinfo{pages}{1318--1330} (\bibinfo{year}{2019}).

\bibitem{bonazzola2024unsupervised}
\bibinfo{author}{Bonazzola, R.} \emph{et~al.}
\newblock \bibinfo{journal}{\bibinfo{title}{Unsupervised ensemble-based phenotyping enhances discoverability of genes related to left-ventricular morphology}}.
\newblock {\emph{\JournalTitle{Nature Machine Intelligence}}} \textbf{\bibinfo{volume}{6}}, \bibinfo{pages}{291--306} (\bibinfo{year}{2024}).

\bibitem{Meyer2020}
\bibinfo{author}{Meyer, H.~V.} \emph{et~al.}
\newblock \bibinfo{journal}{\bibinfo{title}{Genetic and functional insights into the fractal structure of the heart}}.
\newblock {\emph{\JournalTitle{Nature}}} \textbf{\bibinfo{volume}{584}}, \bibinfo{pages}{589--594} (\bibinfo{year}{2020}).

\bibitem{Kim2018}
\bibinfo{author}{Kim, G.~H.}, \bibinfo{author}{Uriel, N.} \& \bibinfo{author}{Burkhoff, D.}
\newblock \bibinfo{journal}{\bibinfo{title}{Reverse remodelling and myocardial recovery in heart failure}}.
\newblock {\emph{\JournalTitle{Nature Reviews Cardiology}}} \textbf{\bibinfo{volume}{15}}, \bibinfo{pages}{83--96} (\bibinfo{year}{2018}).

\bibitem{DBLP:journals/natmi/BelloDDBMHGWCRO19}
\bibinfo{author}{Bello, G.~A.} \emph{et~al.}
\newblock \bibinfo{journal}{\bibinfo{title}{Deep-learning cardiac motion analysis for human survival prediction}}.
\newblock {\emph{\JournalTitle{Nature Machine Intelligence}}}  (\bibinfo{year}{2019}).

\bibitem{puyol2017multimodal}
\bibinfo{author}{Puyol-Anton, E.} \emph{et~al.}
\newblock \bibinfo{journal}{\bibinfo{title}{{A multimodal spatiotemporal cardiac motion atlas from MR and ultrasound data}}}.
\newblock {\emph{\JournalTitle{Medical Image Analysis}}}  (\bibinfo{year}{2017}).

\bibitem{Duchateau2020}
\bibinfo{author}{Duchateau, N.}, \bibinfo{author}{King, A.~P.} \& \bibinfo{author}{De~Craene, M.}
\newblock \bibinfo{journal}{\bibinfo{title}{Machine learning approaches for myocardial motion and deformation analysis}}.
\newblock {\emph{\JournalTitle{Frontiers in Cardiovascular Medicine}}}  (\bibinfo{year}{2020}).

\bibitem{bai2020population}
\bibinfo{author}{Bai, W.} \emph{et~al.}
\newblock \bibinfo{journal}{\bibinfo{title}{A population-based phenome-wide association study of cardiac and aortic structure and function}}.
\newblock {\emph{\JournalTitle{Nature Medicine}}}  (\bibinfo{year}{2020}).

\bibitem{corral2020digital}
\bibinfo{author}{Corral-Acero, J.} \emph{et~al.}
\newblock \bibinfo{journal}{\bibinfo{title}{{The ‘Digital Twin’ to enable the vision of precision cardiology}}}.
\newblock {\emph{\JournalTitle{European Heart Journal}}} \textbf{\bibinfo{volume}{41}}, \bibinfo{pages}{4556--4564} (\bibinfo{year}{2020}).

\bibitem{schulz2020standardized}
\bibinfo{author}{Schulz-Menger, J.} \emph{et~al.}
\newblock \bibinfo{journal}{\bibinfo{title}{Standardized image interpretation and post-processing in cardiovascular magnetic resonance-2020 update: Society for cardiovascular magnetic resonance (scmr): Board of trustees task force on standardized post-processing}}.
\newblock {\emph{\JournalTitle{Journal of Cardiovascular Magnetic Resonance}}} \textbf{\bibinfo{volume}{22}}, \bibinfo{pages}{19} (\bibinfo{year}{2020}).

\bibitem{hundley2022society}
\bibinfo{author}{Hundley, W.~G.} \emph{et~al.}
\newblock \bibinfo{journal}{\bibinfo{title}{{Society for Cardiovascular Magnetic Resonance (SCMR) guidelines for reporting cardiovascular magnetic resonance examinations}}}.
\newblock {\emph{\JournalTitle{Journal of Cardiovascular Magnetic Resonance}}} \textbf{\bibinfo{volume}{24}}, \bibinfo{pages}{29} (\bibinfo{year}{2022}).

\bibitem{gasparovici2024generative}
\bibinfo{author}{Gasparovici, A.} \& \bibinfo{author}{Serban, A.}
\newblock \bibinfo{journal}{\bibinfo{title}{Generative 3d cardiac shape modelling for in-silico trials}}.
\newblock {\emph{\JournalTitle{Studies in health technology and informatics}}} \textbf{\bibinfo{volume}{321}}, \bibinfo{pages}{190--194} (\bibinfo{year}{2024}).

\bibitem{piras2017morphologically}
\bibinfo{author}{Piras, P.} \emph{et~al.}
\newblock \bibinfo{journal}{\bibinfo{title}{Morphologically normalized left ventricular motion indicators from mri feature tracking characterize myocardial infarction}}.
\newblock {\emph{\JournalTitle{Scientific reports}}} \textbf{\bibinfo{volume}{7}}, \bibinfo{pages}{12259} (\bibinfo{year}{2017}).

\bibitem{gilbert2019independent}
\bibinfo{author}{Gilbert, K.} \emph{et~al.}
\newblock \bibinfo{journal}{\bibinfo{title}{Independent left ventricular morphometric atlases show consistent relationships with cardiovascular risk factors: a uk biobank study}}.
\newblock {\emph{\JournalTitle{Scientific reports}}} \textbf{\bibinfo{volume}{9}}, \bibinfo{pages}{1130} (\bibinfo{year}{2019}).

\bibitem{mauger2022multi}
\bibinfo{author}{Mauger, C.~A.} \emph{et~al.}
\newblock \bibinfo{journal}{\bibinfo{title}{Multi-ethnic study of atherosclerosis: relationship between left ventricular shape at cardiac mri and 10-year outcomes}}.
\newblock {\emph{\JournalTitle{Radiology}}} \textbf{\bibinfo{volume}{306}}, \bibinfo{pages}{e220122} (\bibinfo{year}{2022}).

\bibitem{qi2020non}
\bibinfo{author}{Qi, H.} \emph{et~al.}
\newblock \bibinfo{journal}{\bibinfo{title}{Non-rigid respiratory motion estimation of whole-heart coronary mr images using unsupervised deep learning}}.
\newblock {\emph{\JournalTitle{IEEE Transactions on Medical Imaging}}} \textbf{\bibinfo{volume}{40}}, \bibinfo{pages}{444--454} (\bibinfo{year}{2020}).

\bibitem{ye2023sequencemorph}
\bibinfo{author}{Ye, M.} \emph{et~al.}
\newblock \bibinfo{journal}{\bibinfo{title}{Sequencemorph: A unified unsupervised learning framework for motion tracking on cardiac image sequences}}.
\newblock {\emph{\JournalTitle{IEEE Transactions on Pattern Analysis and Machine Intelligence}}}  (\bibinfo{year}{2023}).

\bibitem{Suinesiaputra2017}
\bibinfo{author}{Suinesiaputra, A.} \emph{et~al.}
\newblock \bibinfo{journal}{\bibinfo{title}{{Statistical shape modeling of the left ventricle: Myocardial infarct classification challenge}}}.
\newblock {\emph{\JournalTitle{IEEE Journal of Biomedical and Health Informatics}}}  (\bibinfo{year}{2017}).

\bibitem{Zheng2019}
\bibinfo{author}{Zheng, Q.}, \bibinfo{author}{Delingette, H.} \& \bibinfo{author}{Ayache, N.}
\newblock \bibinfo{journal}{\bibinfo{title}{{Explainable cardiac pathology classification on cine MRI with motion characterization by semi-supervised learning of apparent flow}}}.
\newblock {\emph{\JournalTitle{Medical Image Analysis}}}  (\bibinfo{year}{2019}).

\bibitem{Kawel2020}
\bibinfo{author}{Kawel-Boehm, N.} \emph{et~al.}
\newblock \bibinfo{journal}{\bibinfo{title}{Reference ranges (“normal values”) for cardiovascular magnetic resonance (cmr) in adults and children: 2020 update}}.
\newblock {\emph{\JournalTitle{Journal of Cardiovascular Magnetic Resonance}}} \textbf{\bibinfo{volume}{22}}, \bibinfo{pages}{1--63} (\bibinfo{year}{2020}).

\bibitem{gao2022get3d}
\bibinfo{author}{Gao, J.} \emph{et~al.}
\newblock \bibinfo{journal}{\bibinfo{title}{Get3d: A generative model of high quality 3d textured shapes learned from images}}.
\newblock {\emph{\JournalTitle{Neural Information Processing Systems}}} \textbf{\bibinfo{volume}{35}}, \bibinfo{pages}{31841--31854} (\bibinfo{year}{2022}).

\bibitem{xue2022giraffe}
\bibinfo{author}{Xue, Y.}, \bibinfo{author}{Li, Y.}, \bibinfo{author}{Singh, K.~K.} \& \bibinfo{author}{Lee, Y.~J.}
\newblock \bibinfo{title}{Giraffe hd: A high-resolution 3d-aware generative model}.
\newblock In \emph{\bibinfo{booktitle}{Proceedings of the IEEE/CVF Conference on Computer Vision and Pattern Recognition}}, \bibinfo{pages}{18440--18449} (\bibinfo{year}{2022}).

\bibitem{kim2023datid}
\bibinfo{author}{Kim, G.} \& \bibinfo{author}{Chun, S.~Y.}
\newblock \bibinfo{title}{Datid-3d: Diversity-preserved domain adaptation using text-to-image diffusion for 3d generative model}.
\newblock In \emph{\bibinfo{booktitle}{Proceedings of the IEEE/CVF Conference on Computer Vision and Pattern Recognition}}, \bibinfo{pages}{14203--14213} (\bibinfo{year}{2023}).

\bibitem{DBLP:conf/eccv/PetrovichBV22}
\bibinfo{author}{Petrovich, M.}, \bibinfo{author}{Black, M.~J.} \& \bibinfo{author}{Varol, G.}
\newblock \bibinfo{title}{{TEMOS: Generating diverse human motions from textual descriptions}}.
\newblock In \emph{\bibinfo{booktitle}{European Conference on Computer Vision}} (\bibinfo{year}{2022}).

\bibitem{DBLP:journals/corr/abs-2209-04066}
\bibinfo{author}{Athanasiou, N.}, \bibinfo{author}{Petrovich, M.}, \bibinfo{author}{Black, M.~J.} \& \bibinfo{author}{Varol, G.}
\newblock \bibinfo{title}{{TEACH: Temporal action composition for 3D humans}}.
\newblock In \emph{\bibinfo{booktitle}{International Conference on 3D Vision}} (\bibinfo{year}{2022}).

\bibitem{DBLP:conf/iccv/PetrovichBV21}
\bibinfo{author}{Petrovich, M.}, \bibinfo{author}{Black, M.~J.} \& \bibinfo{author}{Varol, G.}
\newblock \bibinfo{title}{{Action-conditioned 3D human motion synthesis with Transformer VAE}}.
\newblock In \emph{\bibinfo{booktitle}{International Conference on Computer Vision}} (\bibinfo{year}{2021}).

\bibitem{swanson2024generative}
\bibinfo{author}{Swanson, K.} \emph{et~al.}
\newblock \bibinfo{journal}{\bibinfo{title}{{Generative AI for designing and validating easily synthesizable and structurally novel antibiotics}}}.
\newblock {\emph{\JournalTitle{Nature Machine Intelligence}}} \textbf{\bibinfo{volume}{6}}, \bibinfo{pages}{338--353} (\bibinfo{year}{2024}).

\bibitem{jiang2024pocketflow}
\bibinfo{author}{Jiang, Y.} \emph{et~al.}
\newblock \bibinfo{journal}{\bibinfo{title}{Pocketflow is a data-and-knowledge-driven structure-based molecular generative model}}.
\newblock {\emph{\JournalTitle{Nature Machine Intelligence}}} \bibinfo{pages}{1--12} (\bibinfo{year}{2024}).

\bibitem{kong2024sdf4chd}
\bibinfo{author}{Kong, F.} \emph{et~al.}
\newblock \bibinfo{journal}{\bibinfo{title}{Sdf4chd: Generative modeling of cardiac anatomies with congenital heart defects}}.
\newblock {\emph{\JournalTitle{Medical Image Analysis}}} \textbf{\bibinfo{volume}{97}}, \bibinfo{pages}{103293} (\bibinfo{year}{2024}).

\bibitem{wang2020deep}
\bibinfo{author}{Wang, S.} \emph{et~al.}
\newblock \bibinfo{title}{Deep generative model-based quality control for cardiac mri segmentation}.
\newblock In \emph{\bibinfo{booktitle}{Medical Image Computing and Computer Assisted Intervention--MICCAI 2020: 23rd International Conference, Lima, Peru, October 4--8, 2020, Proceedings, Part IV 23}}, \bibinfo{pages}{88--97} (\bibinfo{organization}{Springer}, \bibinfo{year}{2020}).

\bibitem{vukadinovic2023gancmri}
\bibinfo{author}{Vukadinovic, M.}, \bibinfo{author}{Kwan, A.~C.}, \bibinfo{author}{Li, D.} \& \bibinfo{author}{Ouyang, D.}
\newblock \bibinfo{title}{Gancmri: Cardiac magnetic resonance video generation and physiologic guidance using latent space prompting}.
\newblock In \emph{\bibinfo{booktitle}{Machine Learning for Health (ML4H)}}, \bibinfo{pages}{594--606} (\bibinfo{organization}{PMLR}, \bibinfo{year}{2023}).

\bibitem{gomez2022digital}
\bibinfo{author}{G{\'o}mez, S.}, \bibinfo{author}{Romo-Bucheli, D.} \& \bibinfo{author}{Mart{\'\i}nez, F.}
\newblock \bibinfo{journal}{\bibinfo{title}{A digital cardiac disease biomarker from a generative progressive cardiac cine-mri representation}}.
\newblock {\emph{\JournalTitle{Biomedical Engineering Letters}}} \textbf{\bibinfo{volume}{12}}, \bibinfo{pages}{75--84} (\bibinfo{year}{2022}).

\bibitem{muffoletto2023combining}
\bibinfo{author}{Muffoletto, M.} \emph{et~al.}
\newblock \bibinfo{journal}{\bibinfo{title}{Combining generative modelling and semi-supervised domain adaptation for whole heart cardiovascular magnetic resonance angiography segmentation}}.
\newblock {\emph{\JournalTitle{Journal of Cardiovascular Magnetic Resonance}}} \textbf{\bibinfo{volume}{25}}, \bibinfo{pages}{80} (\bibinfo{year}{2023}).

\bibitem{xia2022automatic}
\bibinfo{author}{Xia, Y.} \emph{et~al.}
\newblock \bibinfo{journal}{\bibinfo{title}{{Automatic 3D+t four-chamber CMR quantification of the UK biobank: integrating imaging and non-imaging data priors at scale}}}.
\newblock {\emph{\JournalTitle{Medical Image Analysis}}}  (\bibinfo{year}{2022}).

\bibitem{gaggion2023multi}
\bibinfo{author}{Gaggion, N.} \emph{et~al.}
\newblock \bibinfo{journal}{\bibinfo{title}{{Multi-view hybrid graph convolutional network for volume-to-mesh reconstruction in cardiovascular MRI}}}.
\newblock {\emph{\JournalTitle{arXiv preprint arXiv:2311.13706}}}  (\bibinfo{year}{2023}).

\bibitem{dou2023conditional}
\bibinfo{author}{Dou, H.}, \bibinfo{author}{Ravikumar, N.} \& \bibinfo{author}{Frangi, A.~F.}
\newblock \bibinfo{title}{A conditional flow variational autoencoder for controllable synthesis of virtual populations of anatomy}.
\newblock In \emph{\bibinfo{booktitle}{International Conference on Medical Image Computing and Computer-Assisted Intervention}} (\bibinfo{year}{2023}).

\bibitem{dou2024generative}
\bibinfo{author}{Dou, H.}, \bibinfo{author}{Virtanen, S.}, \bibinfo{author}{Ravikumar, N.} \& \bibinfo{author}{Frangi, A.~F.}
\newblock \bibinfo{journal}{\bibinfo{title}{A generative shape compositional framework to synthesize populations of virtual chimeras}}.
\newblock {\emph{\JournalTitle{IEEE Transactions on Neural Networks and Learning Systems}}}  (\bibinfo{year}{2024}).

\bibitem{beetz2022interpretable}
\bibinfo{author}{Beetz, M.} \emph{et~al.}
\newblock \bibinfo{journal}{\bibinfo{title}{Interpretable cardiac anatomy modeling using variational mesh autoencoders}}.
\newblock {\emph{\JournalTitle{Frontiers in Cardiovascular Medicine}}}  (\bibinfo{year}{2022}).

\bibitem{beetz2021generating}
\bibinfo{author}{Beetz, M.}, \bibinfo{author}{Banerjee, A.} \& \bibinfo{author}{Grau, V.}
\newblock \bibinfo{title}{Generating subpopulation-specific biventricular anatomy models using conditional point cloud variational autoencoders}.
\newblock In \emph{\bibinfo{booktitle}{International Workshop on Statistical Atlases and Computational Models of the Heart}} (\bibinfo{year}{2021}).

\bibitem{campello2022cardiac}
\bibinfo{author}{Campello, V.~M.} \emph{et~al.}
\newblock \bibinfo{journal}{\bibinfo{title}{Cardiac aging synthesis from cross-sectional data with conditional generative adversarial networks}}.
\newblock {\emph{\JournalTitle{Frontiers in Cardiovascular Medicine}}}  (\bibinfo{year}{2022}).

\bibitem{qiao2023cheart}
\bibinfo{author}{Qiao, M.} \emph{et~al.}
\newblock \bibinfo{journal}{\bibinfo{title}{{CHeart: A conditional spatio-temporal generative model for cardiac anatomy}}}.
\newblock {\emph{\JournalTitle{IEEE Transactions on Medical Imaging}}} \textbf{\bibinfo{volume}{43}} (\bibinfo{year}{2024}).

\bibitem{Hadrien2022}
\bibinfo{author}{Reynaud, H.} \emph{et~al.}
\newblock \bibinfo{title}{{D'ARTAGNAN: Counterfactual video generation}}.
\newblock In \emph{\bibinfo{booktitle}{Medical Image Computing and Computer Assisted Intervention}} (\bibinfo{year}{2022}).

\bibitem{Gilbert2021}
\bibinfo{author}{Gilbert, A.} \emph{et~al.}
\newblock \bibinfo{journal}{\bibinfo{title}{{Generating synthetic labeled data from existing anatomical models: An example with echocardiography segmentation}}}.
\newblock {\emph{\JournalTitle{IEEE Transactions on Medical Imaging}}}  (\bibinfo{year}{2021}).

\bibitem{DBLP:conf/iclr/KipfW17}
\bibinfo{author}{Kipf, T.~N.} \& \bibinfo{author}{Welling, M.}
\newblock \bibinfo{title}{Semi-supervised classification with graph convolutional networks}.
\newblock In \emph{\bibinfo{booktitle}{International Conference on Learning Representations}} (\bibinfo{year}{2017}).

\bibitem{petersen2015uk}
\bibinfo{author}{Petersen, S.~E.} \emph{et~al.}
\newblock \bibinfo{journal}{\bibinfo{title}{{UK Biobank’s cardiovascular magnetic resonance protocol}}}.
\newblock {\emph{\JournalTitle{Journal of Cardiovascular Magnetic Resonance}}}  (\bibinfo{year}{2015}).

\bibitem{DBLP:conf/mm/GuoZWZSDG020}
\bibinfo{author}{Guo, C.} \emph{et~al.}
\newblock \bibinfo{title}{{Action2Motion: Conditioned generation of 3D human motions}}.
\newblock In \emph{\bibinfo{booktitle}{{ACM} International Conference on Multimedia}} (\bibinfo{year}{2020}).

\bibitem{cover1999elements}
\bibinfo{author}{Cover, T.~M.}
\newblock \emph{\bibinfo{title}{Elements of information theory}} (\bibinfo{publisher}{John Wiley \& Sons}, \bibinfo{year}{1999}).

\bibitem{arjovsky2017wasserstein}
\bibinfo{author}{Arjovsky, M.}, \bibinfo{author}{Chintala, S.} \& \bibinfo{author}{Bottou, L.}
\newblock \bibinfo{title}{Wasserstein generative adversarial networks}.
\newblock In \emph{\bibinfo{booktitle}{International Conference on Machine Learning}}, \bibinfo{pages}{214--223} (\bibinfo{organization}{PMLR}, \bibinfo{year}{2017}).

\bibitem{van2008visualizing}
\bibinfo{author}{Van~der Maaten, L.} \& \bibinfo{author}{Hinton, G.}
\newblock \bibinfo{journal}{\bibinfo{title}{{Visualizing data using t-SNE}}}.
\newblock {\emph{\JournalTitle{Journal of Machine Learning Research}}} \textbf{\bibinfo{volume}{9}} (\bibinfo{year}{2008}).

\bibitem{fukuta2008cardiac}
\bibinfo{author}{Fukuta, H.} \& \bibinfo{author}{Little, W.~C.}
\newblock \bibinfo{journal}{\bibinfo{title}{The cardiac cycle and the physiologic basis of left ventricular contraction, ejection, relaxation, and filling}}.
\newblock {\emph{\JournalTitle{Heart failure clinics}}} \textbf{\bibinfo{volume}{4}}, \bibinfo{pages}{1--11} (\bibinfo{year}{2008}).

\bibitem{nethononda2015gender}
\bibinfo{author}{Nethononda, R.~M.} \emph{et~al.}
\newblock \bibinfo{journal}{\bibinfo{title}{Gender specific patterns of age-related decline in aortic stiffness: a cardiovascular magnetic resonance study including normal ranges}}.
\newblock {\emph{\JournalTitle{Journal of Cardiovascular Magnetic Resonance}}} \textbf{\bibinfo{volume}{17}}, \bibinfo{pages}{20} (\bibinfo{year}{2015}).

\bibitem{heckbert2006traditional}
\bibinfo{author}{Heckbert, S.~R.} \emph{et~al.}
\newblock \bibinfo{journal}{\bibinfo{title}{Traditional cardiovascular risk factors in relation to left ventricular mass, volume, and systolic function by cardiac magnetic resonance imaging: the multiethnic study of atherosclerosis}}.
\newblock {\emph{\JournalTitle{Journal of the American College of Cardiology}}} \textbf{\bibinfo{volume}{48}}, \bibinfo{pages}{2285--2292} (\bibinfo{year}{2006}).

\bibitem{ortega2016obesity}
\bibinfo{author}{Ortega, F.~B.}, \bibinfo{author}{Lavie, C.~J.} \& \bibinfo{author}{Blair, S.~N.}
\newblock \bibinfo{journal}{\bibinfo{title}{Obesity and cardiovascular disease}}.
\newblock {\emph{\JournalTitle{Circulation Research}}}  (\bibinfo{year}{2016}).

\bibitem{ormazabal2018association}
\bibinfo{author}{Ormazabal, V.} \emph{et~al.}
\newblock \bibinfo{journal}{\bibinfo{title}{Association between insulin resistance and the development of cardiovascular disease}}.
\newblock {\emph{\JournalTitle{Cardiovascular Diabetology}}}  (\bibinfo{year}{2018}).

\bibitem{ford2012ideal}
\bibinfo{author}{Ford, E.~S.}, \bibinfo{author}{Greenlund, K.~J.} \& \bibinfo{author}{Hong, Y.}
\newblock \bibinfo{journal}{\bibinfo{title}{Ideal cardiovascular health and mortality from all causes and diseases of the circulatory system among adults in the united states}}.
\newblock {\emph{\JournalTitle{Circulation}}}  (\bibinfo{year}{2012}).

\bibitem{binu2017heart}
\bibinfo{author}{Binu, A.~J.} \emph{et~al.}
\newblock \bibinfo{journal}{\bibinfo{title}{The heart of the matter: cardiac manifestations of endocrine disease}}.
\newblock {\emph{\JournalTitle{Indian Journal of Endocrinology and Metabolism}}}  (\bibinfo{year}{2017}).

\bibitem{bhatnagar2017environmental}
\bibinfo{author}{Bhatnagar, A.}
\newblock \bibinfo{journal}{\bibinfo{title}{Environmental determinants of cardiovascular disease}}.
\newblock {\emph{\JournalTitle{Circulation research}}} \textbf{\bibinfo{volume}{121}}, \bibinfo{pages}{162--180} (\bibinfo{year}{2017}).

\bibitem{mann2005mechanisms}
\bibinfo{author}{Mann, D.~L.} \& \bibinfo{author}{Bristow, M.~R.}
\newblock \bibinfo{journal}{\bibinfo{title}{Mechanisms and models in heart failure: the biomechanical model and beyond}}.
\newblock {\emph{\JournalTitle{Circulation}}}  (\bibinfo{year}{2005}).

\bibitem{trayanova2011whole}
\bibinfo{author}{Trayanova, N.~A.}
\newblock \bibinfo{journal}{\bibinfo{title}{Whole-heart modeling: applications to cardiac electrophysiology and electromechanics}}.
\newblock {\emph{\JournalTitle{Circulation Research}}}  (\bibinfo{year}{2011}).

\bibitem{mauger2019right}
\bibinfo{author}{Mauger, C.} \emph{et~al.}
\newblock \bibinfo{journal}{\bibinfo{title}{Right ventricular shape and function: cardiovascular magnetic resonance reference morphology and biventricular risk factor morphometrics in uk biobank}}.
\newblock {\emph{\JournalTitle{Journal of Cardiovascular Magnetic Resonance}}} \textbf{\bibinfo{volume}{21}}, \bibinfo{pages}{41} (\bibinfo{year}{2019}).

\bibitem{Dosovitskiy2021}
\bibinfo{author}{Dosovitskiy, A.} \emph{et~al.}
\newblock \bibinfo{title}{{An image is worth 16x16 words: Transformers for image recognition at scale}}.
\newblock In \emph{\bibinfo{booktitle}{International Conference on Learning Representations}} (\bibinfo{year}{2021}).

\bibitem{DBLP:journals/corr/KingmaW13}
\bibinfo{author}{Kingma, D.~P.} \& \bibinfo{author}{Welling, M.}
\newblock \bibinfo{title}{Auto-encoding variational bayes}.
\newblock In \emph{\bibinfo{booktitle}{International Conference on Learning Representations}} (\bibinfo{year}{2014}).

\bibitem{DBLP:conf/iclr/HigginsMPBGBML17}
\bibinfo{author}{Higgins, I.} \emph{et~al.}
\newblock \bibinfo{title}{{beta-VAE: Learning basic visual concepts with a constrained variational framework}}.
\newblock In \emph{\bibinfo{booktitle}{International Conference on Learning Representations}} (\bibinfo{year}{2017}).

\bibitem{fan2017point}
\bibinfo{author}{Fan, H.}, \bibinfo{author}{Su, H.} \& \bibinfo{author}{Guibas, L.~J.}
\newblock \bibinfo{title}{{A point set generation network for 3D object reconstruction from a single image}}.
\newblock In \emph{\bibinfo{booktitle}{IEEE Conference on Computer Vision and Pattern Recognition}} (\bibinfo{year}{2017}).

\bibitem{DBLP:conf/graphite/NealenISA06}
\bibinfo{author}{Nealen, A.}, \bibinfo{author}{Igarashi, T.}, \bibinfo{author}{Sorkine, O.} \& \bibinfo{author}{Alexa, M.}
\newblock \bibinfo{title}{Laplacian mesh optimization}.
\newblock In \emph{\bibinfo{booktitle}{International Conference on Computer Graphics and Interactive Techniques in Australasia and Southeast Asia}}, \bibinfo{pages}{381--389} (\bibinfo{year}{2006}).

\bibitem{desbrun1999implicit}
\bibinfo{author}{Desbrun, M.}, \bibinfo{author}{Meyer, M.}, \bibinfo{author}{Schr{\"o}der, P.} \& \bibinfo{author}{Barr, A.~H.}
\newblock \bibinfo{title}{Implicit fairing of irregular meshes using diffusion and curvature flow}.
\newblock In \emph{\bibinfo{booktitle}{Conference on Computer Graphics and Interactive Techniques}} (\bibinfo{year}{1999}).

\bibitem{bai2018automated}
\bibinfo{author}{Bai, W.} \emph{et~al.}
\newblock \bibinfo{journal}{\bibinfo{title}{Automated cardiovascular magnetic resonance image analysis with fully convolutional networks}}.
\newblock {\emph{\JournalTitle{Journal of cardiovascular magnetic resonance}}} \textbf{\bibinfo{volume}{20}}, \bibinfo{pages}{65} (\bibinfo{year}{2018}).

\bibitem{8624549}
\bibinfo{author}{Duan, J.} \emph{et~al.}
\newblock \bibinfo{journal}{\bibinfo{title}{Automatic 3d bi-ventricular segmentation of cardiac images by a shape-refined multi-task deep learning approach}}.
\newblock {\emph{\JournalTitle{IEEE Transactions on Medical Imaging}}} \textbf{\bibinfo{volume}{38}} (\bibinfo{year}{2019}).

\bibitem{bai2015bi}
\bibinfo{author}{Bai, W.} \emph{et~al.}
\newblock \bibinfo{journal}{\bibinfo{title}{{A bi-ventricular cardiac atlas built from 1000+ high resolution MR images of healthy subjects and an analysis of shape and motion}}}.
\newblock {\emph{\JournalTitle{Medical Image Analysis}}}  (\bibinfo{year}{2015}).

\bibitem{Schuh2024-si}
\bibinfo{author}{Schuh, A.}, \bibinfo{author}{Qiu, H.} \& \bibinfo{author}{{HeartFlow Research}}.
\newblock \bibinfo{title}{deepali: Image, point set, and surface registration in {PyTorch}} (\bibinfo{year}{2024}).

\bibitem{rueckert1999nonrigid}
\bibinfo{author}{Rueckert, D.} \emph{et~al.}
\newblock \bibinfo{journal}{\bibinfo{title}{Nonrigid registration using free-form deformations: application to breast mr images}}.
\newblock {\emph{\JournalTitle{IEEE Transactions on Medical Imaging}}} \textbf{\bibinfo{volume}{18}}, \bibinfo{pages}{712--721} (\bibinfo{year}{1999}).

\end{thebibliography}
\end{document}